%% file: iclr2021_conference.tex
\documentclass{article} % For LaTeX2e
\usepackage{iclr2021_conference,times}

\input{math_commands.tex}

\usepackage{hyperref}
\usepackage{url}
\usepackage{graphicx}
\usepackage{adjustbox}
\usepackage{epstopdf} %%package to overcome problem with eps in pdf files

\title{MiniGPT-Reverse-Designing: Predicting Image Adjustments Utilizing MiniGPT-4}

\author{Vahid Azizi, Fatemeh Koochaki\\
\texttt{\{va.azizi, fatmakoochaki\}@gmail.com} \\
}

\iclrfinalcopy % Uncomment for camera-ready version, but NOT for submission.
\begin{document}

\maketitle

\begin{abstract}
Vision-Language Models (VLMs) have recently seen significant advancements through integrating with Large Language Models (LLMs). The VLMs, which process image and text modalities simultaneously, have demonstrated the ability to learn and understand the interaction between images and texts across various multi-modal tasks. Reverse designing, which could be defined as a complex vision-language task, aims to predict the edits and their parameters, given a source image, an edited version, and an optional high-level textual edit description. This task requires VLMs to comprehend the interplay between the source image, the edited version, and the optional textual context simultaneously, going beyond traditional vision-language tasks. In this paper, we extend and fine-tune MiniGPT-4 for the reverse designing task. Our experiments demonstrate the extensibility of off-the-shelf VLMs, specifically MiniGPT-4, for more complex tasks such as reverse designing. Code is available at this 
\href{https://github.com/VahidAz/MiniGPT-Reverse-Designing}{repository}.

\end{abstract}

\section{Introduction}
Language Models (LMs) and, recently, Large Language Models (LLMs) are primarily trained on and applied to text data for tasks such as text encoding and generation. However, they can not understand and process other types of data. The frequent co-occurrence of image and text data has driven the development of Vision Language Models (VLMs). These models can understand texts and images simultaneously, enabling them to perform sophisticated multi-modal tasks such as Visual Question Answering (VQA), image captioning, text-to-image search, and multi-modal generation. Pretrained VLMs reduce the computational expense associated with training models from scratch, leading to the emergence of novel applications \cite{ zhou2023vision+, yin2024survey, caffagni2024revolution, 10386743, xu2023multimodal, zhao2023survey, minaee2024large, 10.1145/3641289, 10.1145/3649449, zhang2024visionlanguage}.

% These models can process and understand language (text) and vision (image) simultaneously, 

% (L)VLMs combine images with corresponding texts to create meaningful links between visual and linguistic information. This integration allows them to achieve a deeper comprehension and interaction with multimedia content, enabling them to execute sophisticated vision-language tasks . \cite{baltrušaitis2017multimodal, yin2024survey, caffagni2024revolution, 10386743, 10123038}.

Unlike forward designing methods, which directly apply changes from one image to another, such as in style transfer \cite{7780634, li2017universal} and image-to-image translation \cite{Murez_2018_CVPR}, reverse designing remains relatively unexplored \cite{guhan-etal-2024-tame}. Forward designing replicates the complete transformation from the source image to the target image without understanding or controlling the underlying adjustments. On the other hand, reverse designing aims to extract the underlying edits given a source image, its edited version, and an optional textual description of the edits. Reverse designing is advantageous for various applications, including image editing revision control \cite{rinaldi2023nodegit} and editing visualization \cite{feng2023xnli}. It has been shown that relying solely on source images and edited versions can lead to ambiguity without additional context, such as textual instructions. Using textual descriptions generated during forward designing can provide relevant context for the edits and help to understand the semantic relationships between images \cite{guhan-etal-2024-tame}.

Reverse designing requires understanding how a source image, its edited version, and optional edit description are related, making it a good fit for a complex visual-language task. It has been shown that VLMs can learn meaningful interaction between a pair of images and texts. However, exploring the capability of these models to learn the meaningful interactions between multiple images and a corresponding text is limited. In this work, we explore the extendability of one of the open-source VLMs called MiniGPT-4 \cite{zhu2023minigpt4} and fine-tune it further for reverse-designing (called MiniGPT-Reverse-Designing) and our experiments show that the MiniGPT-4 is extendable to complex multi-modal tasks like reverse designing. 

\section{Method}
\begin{figure}[h]
\centering
\includegraphics[width=\textwidth]{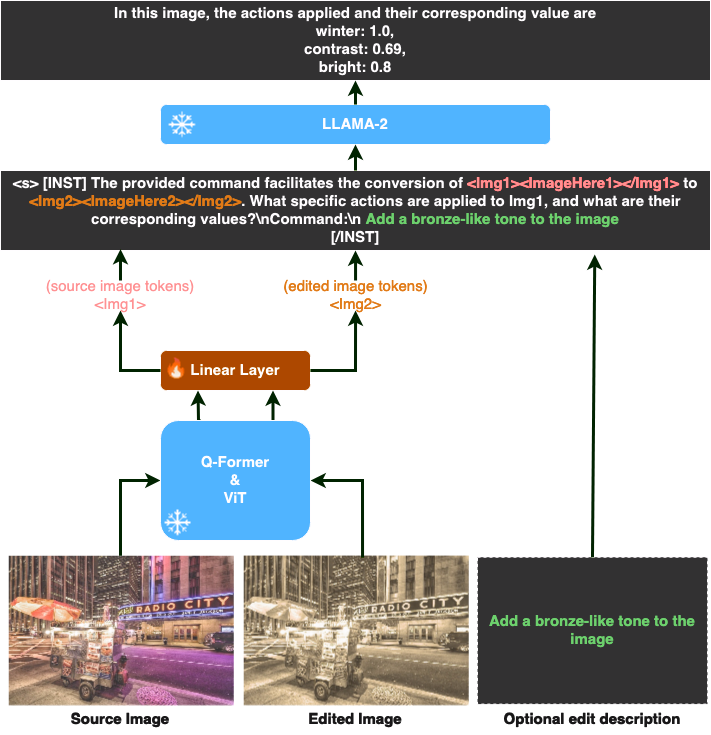}
\caption{MiniGPT-Reverse-Designing architecture. Both the source and edited images are processed through the same pre-trained frozen vision encoder. A learnable linear layer aligns the image features with textual data. The resulting image tokens are then integrated with text tokens using a template prompt. This combined input is fed into the frozen LLAMA-2, which predicts the edit operations and their corresponding values.}
\label{modeldiag}
\end{figure}

In this paper, we explore the extendability of an open-source VLM called MiniGPT-4~\cite{zhu2023minigpt4} for reverse designing to learn the interplay between two images and their optional corresponding text~\cite{guhan-etal-2024-tame}. We focus on predicting the history of edits on an image by giving the source image, its edited version, and an optional high-level description. MiniGPT-4 aligns a frozen visual encoder with a frozen language model by training a projection linear layer \cite{zhu2023minigpt4}. It uses the same visual encoder as BLIP-2 \cite{li2023blip2}, featuring a ViT backbone \cite{dosovitskiy2021image} along with their pre-trained Q-Former \cite{zhu2023minigpt4}. While there are opportunities to modify the MiniGPT-4 architecture or integrate it with additional modules, we aim to explore the capabilities of state-of-the-art architectures for the reverse designing task, and possible changes in architecture are kept for future works. The architecture of MiniGPT-Reverse-Designing is depicted in Fig.~\ref{modeldiag}. Following the original MiniGPT-4 architecture, we only train the linear projection layer to align both source and edited images with the textual context. Both images are converted to tokens through the same visual encoder and aligned with textual context via a learnable linear projection layer. After integrating image and textual tokens using predefined prompts, the inputs are passed to a frozen LLM. MiniGPT-4 used Vicuna \cite{chiang2023vicuna} and LLAMA-2~\cite{touvron2023llama} as frozen LLMs. However, we used only LLAMA-2.

\section{Experiments}
\label{expr}
This section presents the dataset and prompt design used in our study, the experimental setup, and our results and observations across different scenarios. First, we fine-tuned MiniGPT-4 for this task without any modifications. In the second scenario, we introduced auxiliary losses to enhance the model's performance and analyzed its behavior. In the third scenario, we examined the impact of incorporating new tokens on the model's performance.

\subsection{Dataset}
\label{dataset}
\begin{figure}[]
\centering
\includegraphics[width=\textwidth]{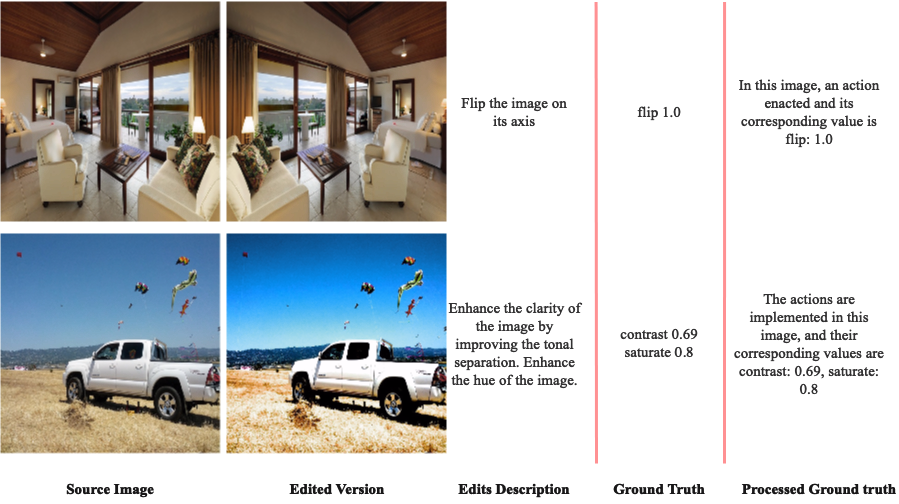}
\caption{The figure presents two dataset examples: one with a single operation and the other with multiple operations. Each sample includes a source image, an edited image, and a high-level edit description. The ground truth, which consists of the operations and their corresponding values, has been converted to sentences as shown in the rightmost column (see Section \ref{setup}).}
\label{fig:datasample}
\end{figure}

% which is derived from the COCO dataset \cite{lin2015microsoft}.
We utilized the Image Multi-Adjustment Dataset (I-MAD)\footnote{\url{https://gamma.umd.edu/researchdirections/affectivecomputing/tame_rd/}} \cite{guhan-etal-2024-tame}, which is available in two versions: I-MAD-Dense and I-MAD-Pro. I-MAD-Dense was generated following the procedure described in \cite{guhan-etal-2024-tame}, whereas I-MAD-Pro is a smaller, human-generated version that offers more accurate samples. For our study, we exclusively used I-MAD-Dense. I-MAD-Dense comprises approximately 22,000 triplets, each containing a source image, an edited image, and a high-level creative editing idea described in the text. The textual descriptions are intentionally vague, omitting specific details about the operations or their parameters. To ensure numerical stability during training and consistent representation across all operations, the values of the operations were normalized \cite{guhan-etal-2024-tame}. Sample triplets are illustrated in Fig. \ref{fig:datasample}. The dataset underwent a preprocessing step to convert the ground truth operations and their values into sentences (see Section \ref{setup} and the rightmost column in Fig. \ref{fig:datasample}).

\subsection{Prompt Design}
\label{prompts}
\begin{figure}[h]
\centering
\includegraphics[width=\textwidth]{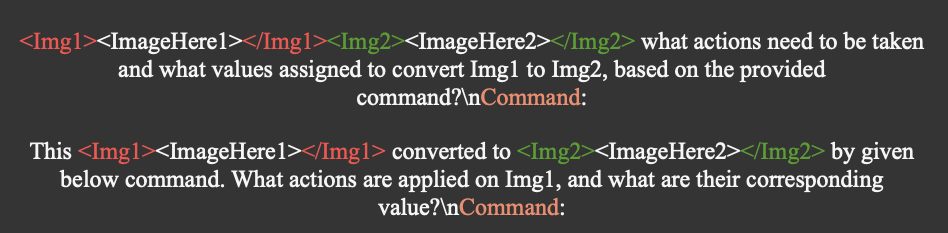}
\caption{Examples of the prompt templates that include the edit description in the form of a command.}
\label{pwcc}
\end{figure}

\begin{figure}[h]
\centering
\includegraphics[width=\textwidth]{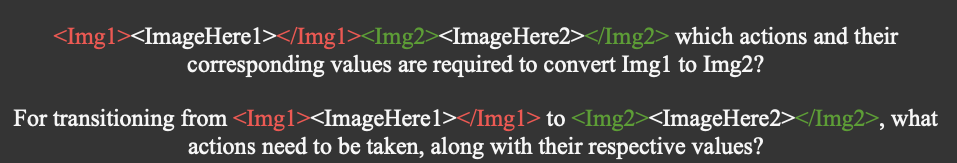}
\caption{Examples of the prompt templates without the inclusion of the edit description.}
\label{pwocc}
\end{figure}

We designed two sets of prompts, each comprising eight templates. One set includes the creative edit idea in the form of a command (Fig.~\ref{pwcc}), while the other set does not (Fig.~\ref{pwocc}). In each set, images are positioned at the beginning in half of the templates, while in the other half, images are placed in arbitrary locations (see Fig.~\ref{pwcc} and Fig.~\ref{pwocc}). It is important to note that we do not calculate the loss for these prompts (see Section~\ref{setup}). All designed prompts are provided in the repository.

\subsection{Experimental Setup}
\label{setup}

\begin{figure}[h]
\centering
\includegraphics[width=\textwidth]{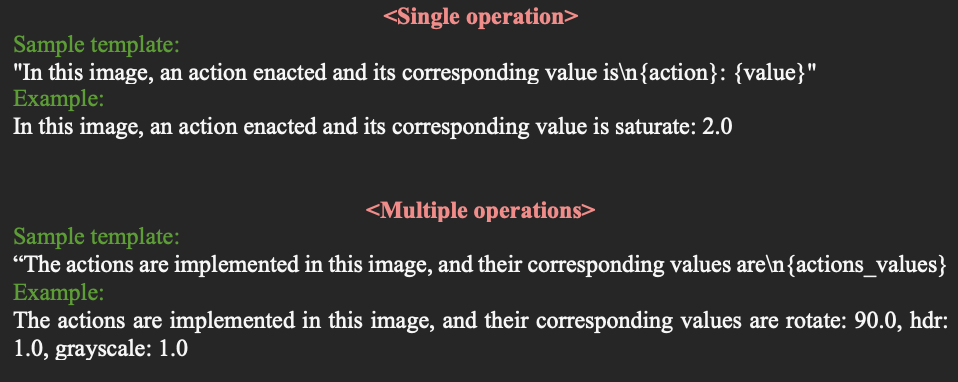}
\caption{Example of templates for converting the ground truth into sentences.}
\label{fig:outtemp}
\end{figure}

\textit{\textbf{Dataset preparation;}} The dataset (see Section \ref{dataset}) contains slightly more than 22,000 records. For our experiments, we randomly sampled 80\% of the data for training and 10\% for validation and used the remaining 10\% for testing and reporting metrics. The split was without overlap; in each set, images are unique. The dataset provides ground truth regarding operations and their corresponding parameters (Fig. \ref{fig:datasample}). To adapt the task to a vision-language task, we converted the ground truth into sentences by sampling from a limited set of designed templates based on the number of operations, as shown in Fig. \ref{fig:outtemp}. To study the effect of textual descriptions, during training, validation, and testing, half of the data in each batch was augmented with the corresponding high-level edit's textual description, randomly sampled from a set of designed prompts (see Section \ref{prompts}). For the other half, we ignored the textual descriptions of the edits, using only images augmented with randomly selected prompts designed for them (see Section \ref{prompts}).

\textit{\textbf{Training configurations;}} In all experiments, we initialized the model with weights obtained from stage-2 fine-tuning of MiniGPT-4 (using LLAMA-2). Due to computational limitations, we restricted the number of experiments, although many more configurations could potentially be tested. To further limit our scope, we kept all training parameters identical to those used in MiniGPT-4 stage-1 training. In earlier experiments, we observed that the model did not perform well with a limited number of iterations. Therefore, in all experiments, we trained the model for ten epochs. The batch size was set to two due to computational constraints. To avoid overfitting, we used the best checkpoint based on validation performance to calculate metrics on the test set. Please note that the loss is not calculated for prompts.

\textit{\textbf{Metrics;}} We observed that the model often overpredicts or underpredicts the number of operations, which prevents us from using standard metrics like average precision (AP) or F1 score. Instead, we report recall by calculating the intersection of predicted operations with the ground truth operations and dividing by the number of ground truth operations. This recall is then averaged over all samples in the test set. To assess the model's performance regarding operation parameters, we report the Mean Square Error (MSE) between the predicted and ground truth parameters. We treat the parameters as zero for operations not present in the predictions. In all experiments, inference was performed using greedy decoding. We also used regex to parse the model output and extract operations and their parameters.

\subsection{Results}

\textit{\textbf{Experiment No. 1: Fine-tuning original MiniGPT-4;}} In the first experiment, we fine-tuned the model without any modifications. The results are provided in Table~\ref{table:results}.

\textit{\textbf{Experiment No. 2: Fine-tuning MiniGPT-4 with MSE as Auxiliary Loss;}} In this experiment, we included the MSE between the predicted operation parameters and the ground truth parameters as an auxiliary loss. Incorporating MSE as an auxiliary loss slightly improved the performance metrics (see Table \ref{table:results}).

\textit{\textbf{Experiment No. 3: Fine-tuning MiniGPT-4 with a Heuristic as Auxiliary Loss;}} To address the issue of underprediction and overprediction, we fine-tuned MiniGPT-4 using a heuristic auxiliary loss comprising three components. The first component penalizes predictions that contain more or fewer operations than the ground truth. The second component penalizes cases where the intersection of predicted and actual operations is less than one. The third component calculates the MSE over the operation parameters. The average of these three components is added to the primary loss. Contrary to our expectations, this auxiliary loss decreased model performance, though MSE (With Command) improved relative to the baseline (see Table \ref{table:results}).

\begin{figure}[h]
\centering
\includegraphics[width=\textwidth,width=0.8\textwidth]{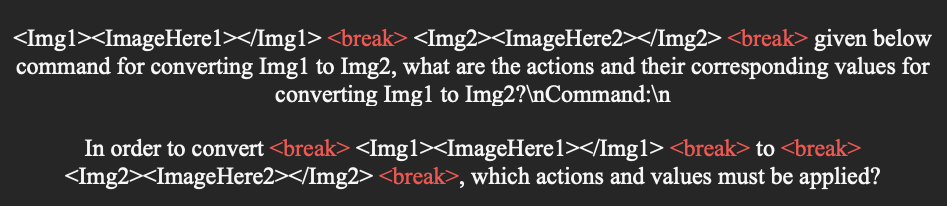}
\caption{Example of prompts using \textit{$\langle break \rangle$} as a special token for transitioning between modalities.}
\label{fig:prombr}
\end{figure}

\textit{\textbf{Experiment No. 4: Fine-Tuning MiniGPT-4 by Adding Special Tokens;}} In \cite{yu2023scaling}, the \textit{$\langle break\rangle$} token is used as an indicator for transitioning between modalities. In MiniGPT-4, the \textit{$\langle img \rangle$} tag is used to represent image tokens, but these tags are not introduced as special tokens and are tokenized during training. Previous experiments followed similar patterns, using tags like \textit{$\langle img1 \rangle$} and \textit{$\langle img2 \rangle$} (see Section \ref{prompts}). Special tokens, introduced during the tokenization process for specific purposes in natural language processing tasks, are not derived from the original text and are not further tokenized. Instead, they provide additional information or perform specific functions, such as the \textit{[SEP]} token in BERT \cite{devlin2019bert} for separating questions and answers. \cite{gao2024enhancing} expanded the vocabulary by adding coordinates as new tokens to aid in predicting correct coordinates. However, adding new tokens generally requires training the LM from scratch. Given the reliance on pre-trained models in MiniGPT-4 and the lack of a large dataset for training from scratch, we could not add operations or operation parameters to the vocabulary. Nevertheless, we conducted two experiments introducing special tokens. The embedding vectors of these added special tokens were averaged over their token embedding vectors. In the first experiment, we added the \textit{$\langle break \rangle$} token to indicate transitions between each entity in the input (see Fig. \ref{fig:prombr}). This approach decreased the metrics overall, with the exception of MSE (With Command), as the results show (see Table \ref{table:results}). We then conducted another experiment by adding more special tokens following the previous explanation, but the performance dropped substantially, and the metrics were not reported. This experiment shows that adding special tokens for this task is not helpful.

\begin{figure}[h]
\centering
\includegraphics[width=\textwidth]{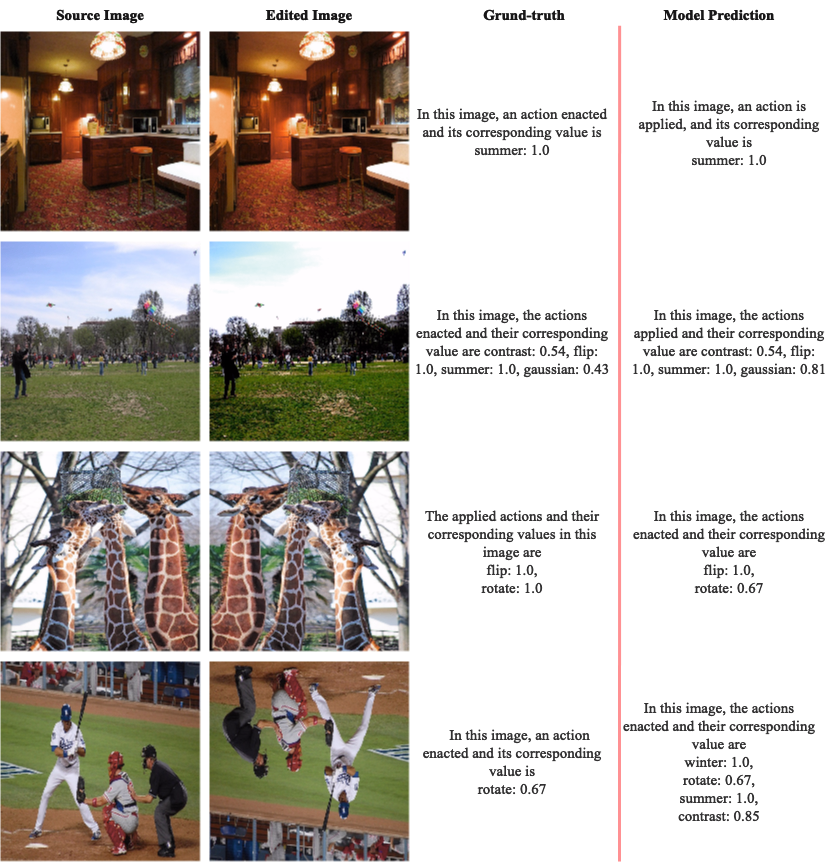}
\caption{Examples of model predictions: The samples in the first three rows include edit descriptions, while the sample in the last row does not.}
\label{fig:ressample}
\end{figure}

\begin{table}[]
\centering
\caption{Evaluation metrics.}
\label{table:results}
\begin{adjustbox}{width=1\textwidth, height=0.065\textwidth}
\begin{tabular}{|c|c|c|c|c|c|c|}
\hline
\textbf{Experiments/Metrics} & \textbf{\begin{tabular}[c]{@{}c@{}}Recall\\ (Average)$\uparrow$\end{tabular}} & \textbf{\begin{tabular}[c]{@{}c@{}}MSE\\ (Average)$\downarrow$\end{tabular}} & \textbf{\begin{tabular}[c]{@{}c@{}}Recall\\ (With Command)$\uparrow$\end{tabular}} & \textbf{\begin{tabular}[c]{@{}c@{}}MSE\\ (With Command)$\downarrow$\end{tabular}} & \textbf{\begin{tabular}[c]{@{}c@{}}Recall\\ (Without Command)$\uparrow$\end{tabular}} & \textbf{\begin{tabular}[c]{@{}c@{}}MSE\\ (Without Command)$\downarrow$\end{tabular}} \\ \hline
\textbf{Experiment No. 1}    & 71.96                                                                 & 0.29                                                             & 93.78                                                                      & 0.17                                                                  & 50.14                                                                         & 0.40                                                                     \\ \hline
\textbf{Experiment No. 2}    & {\color[HTML]{FE0000} \textbf{72.5}}                                    & {\color[HTML]{FE0000} \textbf{0.23}}                             & {\color[HTML]{FE0000} \textbf{94.37}}                                      & {\color[HTML]{FE0000} \textbf{0.09}}                                  & {\color[HTML]{FE0000} \textbf{50.63}}                                         & {\color[HTML]{FE0000} \textbf{0.37}}                                     \\ \hline
\textbf{Experiment No. 3}    & 70.69                                                                 & 0.26                                                             & 92.81                                                                      & 0.11                                                                  & 48.57                                                                         & 0.42                                                                     \\ \hline
\textbf{Experiment No. 4}    & 68.98                                                                 & 0.28                                                             & 92.04                                                                      & 0.11                                                                  & 45.93                                                                         & 0.44                                                                     \\ \hline
\end{tabular}
\end{adjustbox}
\end{table}

\textit{\textbf{Results;}} Experiment No. 2: Fine-tuning MiniGPT-4 with MSE as an auxiliary loss demonstrated superior performance compared to other experiments (see Table \ref{table:results}), furthermore, including high-level textual descriptions of edits notably enhanced performance across all experiments, aligning with previous observations (as noted in \cite{guhan-etal-2024-tame}). We further qualitatively explored the results, and we observed that the model usually performs better for samples with longer edit descriptions (see some samples in Fig.~\ref{fig:ressample}).

\section{Conclusion}
We present MiniGPT-Reverse-Designing, a study exploring the capacity of MiniGPT-4 for the reverse designing task. Reverse designing involves predicting multiple adjustments on a given source image, its edited counterpart, and, optionally, a textual description of the edits. Compared to conventional vision-language tasks, reverse designing poses a more intricate challenge. The model must comprehend the interaction between two images and their accompanying textual descriptions. We fine-tuned the model in various configurations, yielding promising results; however, there remains scope for improvement. Potential enhancements include modifying the model architecture or integrating additional components. Furthermore, leveraging a high-resolution image encoder could augment model performance by capturing finer image details. Another round of fine-tuning with I-MAD-Pro or integrating human feedback in training~\cite{rafailov2024directpreferenceoptimizationlanguage} are other possible directions for improving the performance.



\bibliography{iclr2021_conference}
\bibliographystyle{iclr2021_conference}

% \appendix
% \section{Appendix}

\end{document}

%% file: math_commands.tex
\usepackage{amsmath,amsfonts,bm}

\def\eqref#1{equation~\ref{#1}}
\def\1{\bm{1}}

\DeclareMathAlphabet{\mathsfit}{\encodingdefault}{\sfdefault}{m}{sl}
\SetMathAlphabet{\mathsfit}{bold}{\encodingdefault}{\sfdefault}{bx}{n}